\documentclass[runningheads]{llncs}
\usepackage{amsmath}
\usepackage{amssymb}
\usepackage{booktabs}
\usepackage{comment}
\usepackage{graphicx}
\usepackage{multirow}
\usepackage{tabularx}
\usepackage{algorithm}
\usepackage{algpseudocode} 
\usepackage{subcaption}
\algnewcommand\algorithmicforeach{\textbf{for each}}
\algdef{S}[FOR]{ForEach}[1]{\algorithmicforeach\ #1\ \algorithmicdo}
\begin{document}
\title{CrisisKAN: Knowledge-infused and Explainable Multimodal Attention Network for Crisis Event Classification}

\titlerunning{CrisisKAN}

\author{Shubham Gupta\inst{1}\orcidID{0000-0003-4908-843X} \and
Nandini Saini\inst{1}\orcidID{0000-0003-1736-943X} \and
Suman Kundu\inst{1}\orcidID{0000-0002-7856-4768} \and
Debasis Das\inst{1}\orcidID{0000-0001-6205-4096}}
\authorrunning{Gupta \& Saini et al.}
%
\institute{Department of Computer and Science Engineering, Indian Institute of Technology Jodhpur, India  \\
\email{\{gupta.37, saini.9, suman, debasis\}@iitj.ac.in}}

\maketitle              
\begin{abstract}
Pervasive use of social media has become the emerging source for real-time information (like images, text, or both) to identify various events. Despite the rapid growth of image and text-based event classification, the state-of-the-art (SOTA) models find it challenging to bridge the semantic gap between features of image and text modalities due to inconsistent encoding. Also, the black-box nature of models fails to explain the model's outcomes for building trust in high-stakes situations such as disasters, pandemic. Additionally, the word limit imposed on social media posts can potentially introduce bias towards specific events. To address these issues, we proposed CrisisKAN, a novel \textbf{K}nowledge-infused and Explainable Multimodal \textbf{A}ttention \textbf{N}etwork that entails images and texts in conjunction with external knowledge from Wikipedia to classify crisis events. To enrich the context-specific understanding of textual information, we integrated Wikipedia knowledge using proposed wiki extraction algorithm. Along with this, a guided cross-attention module is implemented to fill the semantic gap in integrating visual and textual data. In order to ensure reliability, we employ a model-specific approach called Gradient-weighted Class Activation Mapping (Grad-CAM) that provides a robust explanation of the predictions of the proposed model. The comprehensive experiments conducted on the CrisisMMD dataset yield in-depth analysis across various crisis-specific tasks and settings. As a result, CrisisKAN outperforms existing SOTA methodologies and provides a novel view in the domain of explainable multimodal event classification.\footnote{Code repository: \url{https://github.com/shubhamgpt007/CrisisKAN} }

\keywords{Multimodal Network \and Explainable \and   Knowledge Infusion \and Crisis Detection.}
\end{abstract}

\section{Introduction}
\label{sec:intro}
With the fast-growing popularity of the internet, social media platforms have become vital medium for the early identification and detection of crisis events such as hurricanes \cite{hunt2020misinformation}, earthquakes \cite{sakaki2010earthquake}, and disease outbreaks like COVID-19 \cite{kwan2020understanding}. These platforms provide massive number of user generated information in the form of either text, image or both for same event within milliseconds. At the same time, rapidly evolving computer vision \cite{he2017mask,mao2022towards} and natural language processing \cite{devlin2018BERT,dai2019transformer} methods are being leveraged to use these data for examining and categorizing crisis events.

\begin{figure}[!t]
    \centering
    \includegraphics[width=.7\linewidth]{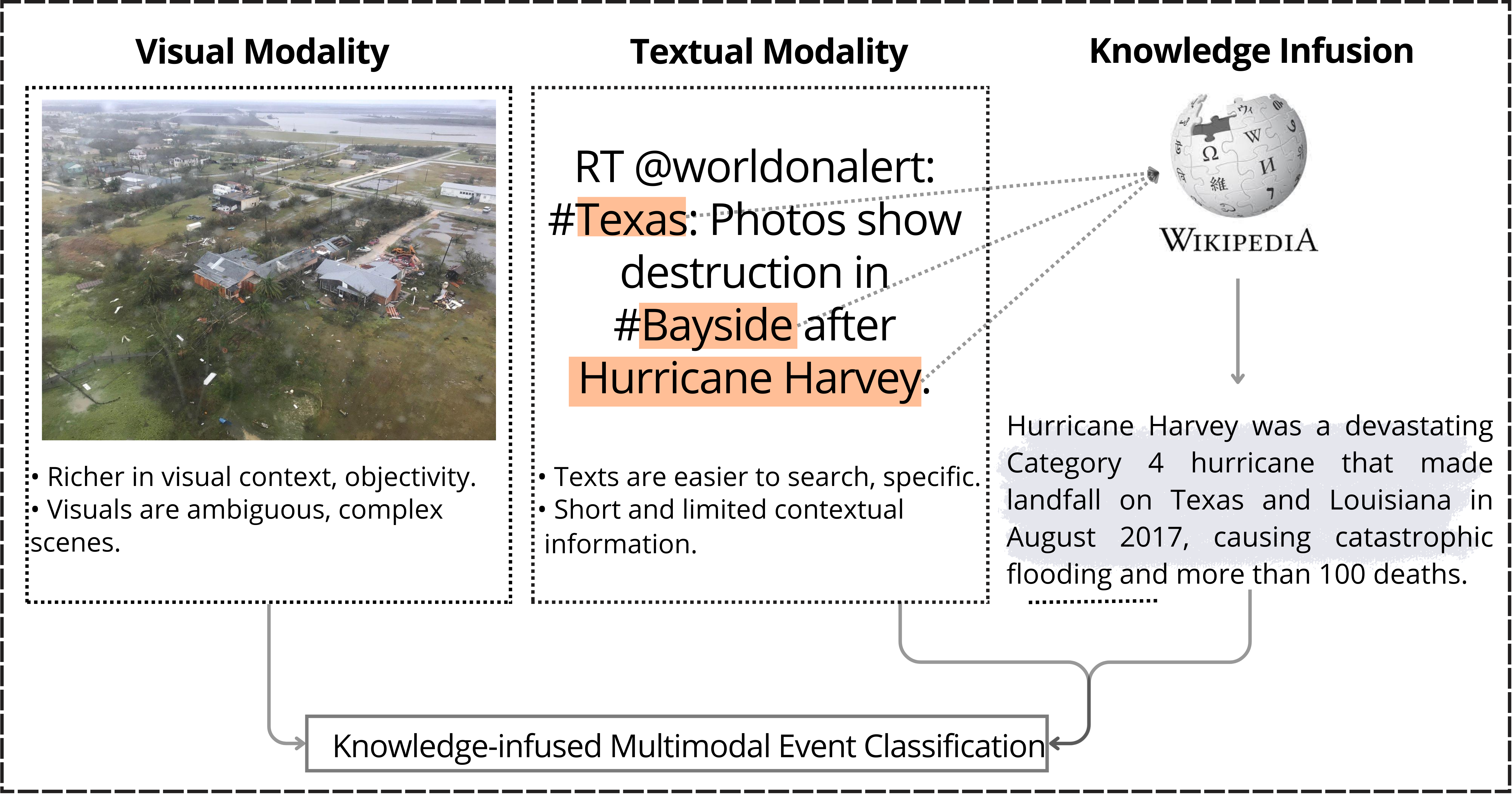}
    \caption{Illustration of Twitter example using knowledge enhanced multimodal event classification to address challenges in visual-textual modality}
    \label{fig:intro}
\end{figure}

Some of the recent works \cite{kiela2018efficient,abavisani2020multimodal,liang2022,singh2022flava} have been proposed  for image-text based multimodal event classification using high-level feature fusion strategy. In these models, feature extraction performed using distinct backbone for each modality, and the aggregation method was utilised for final outcome. However, different aggregation methodologies can create a semantic gap between individual modalities due to inconsistent encoding methods. Further, these backbones are based on deep neural network which can be viewed as black-box and not interpretable. Nevertheless, these methods considered knowledge available from the social media post which is limited due to text length constraint imposed by many social media platforms.


In order to address these issues, we present a novel explainable multimodal framework, named as CrisisKAN (\textbf{K}nowledge-infused and Explainable Multimodal \textbf{A}ttention \textbf{N}etwork), to classify crisis events. CrisisKAN consists a series of phases starting with knowledge infusion followed by image-text feature extraction, multimodal classification, and model explanation. The knowledge infusion step leverages external information by integrating Wikipedia knowledge using a proposed wiki extraction algorithm to enhance the knowledge available with the limited text. Fig.\ref{fig:intro} illustrates an instance of external knowledge extraction through Wikipedia that includes entities information such as `Hurricane Harvey', `Texas', and `Bayside'. Subsequently, feature extraction is applied to both modalities, and their fusion is performed through a guided cross-attention module, that effectively bridges the semantic gap between distinct feature sets. To instill confidence in the model's predictions a model-specific explainable module is also integrated that enables the analysis of feature maps within the black box model. This module also serves as a qualitative performance parameter alongside quantitative (accuracy, F1 etc.). The extensive experiments on crisisMMD \cite{alam2018crisismmd} dataset demonstrate the superior performance of the CrisisKAN compared to existing state-of-the-art solutions, highlighting its effectiveness in crisis event classification. We also propose a new metric Multi-task Model Strength (MTMS) which provides performance of an individual model across different tasks. In summary, our contribution of this work are four fold:
\begin{itemize}

    \item First of all, we exploit knowledge infusion in the multimodal crisis event classification. The proposed implementation effectively overcoming limitations of short text and event biases.

    \item We integrated Guided Cross Attention mechanism to fill the semantic gap among the modalities while aggregating various large pre-trained unimodal classification model. 

    \item Exaplainablity (XAI) of the outcome is incorporated in the model. Proposed XAI layer in the model not only excels on diverse image-text multimodal classification benchmarks but also ensures transparency and interpretability in the outcomes.

    \item We propose a new metric Multi-task Model Strength (MTMS) that provide model performance in more generalization manner across various tasks.
    
\end{itemize}
This paper's overall structure is as follows: In Section \ref{sec:relwork}, we add the study about various classification models and explainability. The proposed methodology is detailed in Section \ref{sec:method}, while Section \ref{sec:exp} describes the experimental setup and discusses the results. Finally, Section \ref{sec:conc} provides a summary of the proposed work and identifies directions for future research.

\section{Related Work}
\label{sec:relwork}
The goal of multimodal learning is focused on integrating different modalities information into a single representation to enrich the representation of data, enabling more robust and accurate predictions. Recent works such as image-text matching \cite{9706873}, multimodal detection \cite{10030494,10030524}, sarcasm detection in memes \cite{bandyopadhyay2023knowledge} and many more take advantage of the diverse information available in both image and text forms. These studies integrate the different modality of features broadly based on three strategies: high-level, intermediate, and low-level feature fusion. In the first strategy, independent deep neural networks are used to generate high-level features for each modality \cite{gallo2020image,Gupta_2023}. Following that, fusion occurs at the model's final layers using aggregation methods such as summation \cite{kiela2018efficient}, tensor fusion \cite{zadeh2017tensor}, OR function based fusion \cite{agarwal2020memis} etc. Wang et al. \cite{wang2021combine} proposed an event detection model that combines low and high-level features to capitalise on their respective advantages and Gupta et al. \cite{GUPTA2023120890} proposed a community based unsupervised event detection model that forms the interaction graph over text modality and apply community detection algorithm to find out micro-level events. On the contrary, intermediate-level feature fusion strategies were optimised that introduce multimodal BERT  by focusing on fine-grained token features of image and text modality \cite{kiela2019supervised,li2020unicoder}. However, these strategies are still limited in their capacity to semantically align features for each modality. Also, there is loss of fine-grained information due to limitation on length in social media text. 

On the other hand, it is evident from the literature that  knowledge infusion methods are advantageous to enhance context-specific understanding to enrich the unimodal representation for better performance. 
Recent research has expanded knowledge by topic wise contextual knowledge \cite{8316358} and social profile information based knowledge \cite{shu2020} for tweet classification task to overcome the short text limitation. Later, researchers \cite{Tahayna_Ayyasamy_Akbar_2022,anonymous2023ean} expand tweet representation through Wikipedia to perform sentiment classification task. This motivates us to explore a novel direction in the field of multimodal crisis event classification.

Further, accuracy may not always a sufficient parameter to evaluate the models for critical applications where explainability is a necessary component. Recently, the eXplainable Artificial Intelligence (XAI) has become a prominent research area to understand the black box functionality of the deep neural network in various domains \cite{adadi2018peeking,petsiuk2021black,moraliyage2022multimodal}.
In a multimodal context, where the model's decision is influenced by the integration of two or more modalities, eXplainable Artificial Intelligence (XAI) plays a crucial role in diagnosing errors. This aids in enhancing model performance, tackling biases, and refining the entire system \cite{joshi2021mmreview,HOLZINGER202128mmXAI}.
The explainability of deep learning models provides justification on predicted outcome of model by means of qualitative or quantitative measures, which can be important for decision-making in high-stakes situation such as disaster, pandemic or medical imaging etc. This shows the importance of explainability in multimodal crisis event classification to effectively explain feature vectors which is responsible in the model's outcome.

\section{Methodology}
\label{sec:method}
The proposed architecture of CrisisKAN is intended to solve the classification problem for crisis events. This section outlines comprehensively the problem definition and sub-module of developed methodology. 

\subsection{Problem Definition}
Given a training dataset $D={\{(I_i,T_i,y_i)\}}_{i=1}^{N}$, where image $I_i$ and text $T_i$ represent a crisis event labeled with $y_i \in C$, the objective is to learn a function $f$ such that $f : f(I_i,T_i) \rightarrow y_i$. Here, $C$ is the set of crisis events. The aim is to enhance predictive accuracy by leveraging information from both modalities.

\subsection{Model Overview}
\begin{figure}[!t]
    \centering
    \includegraphics[width=\linewidth]{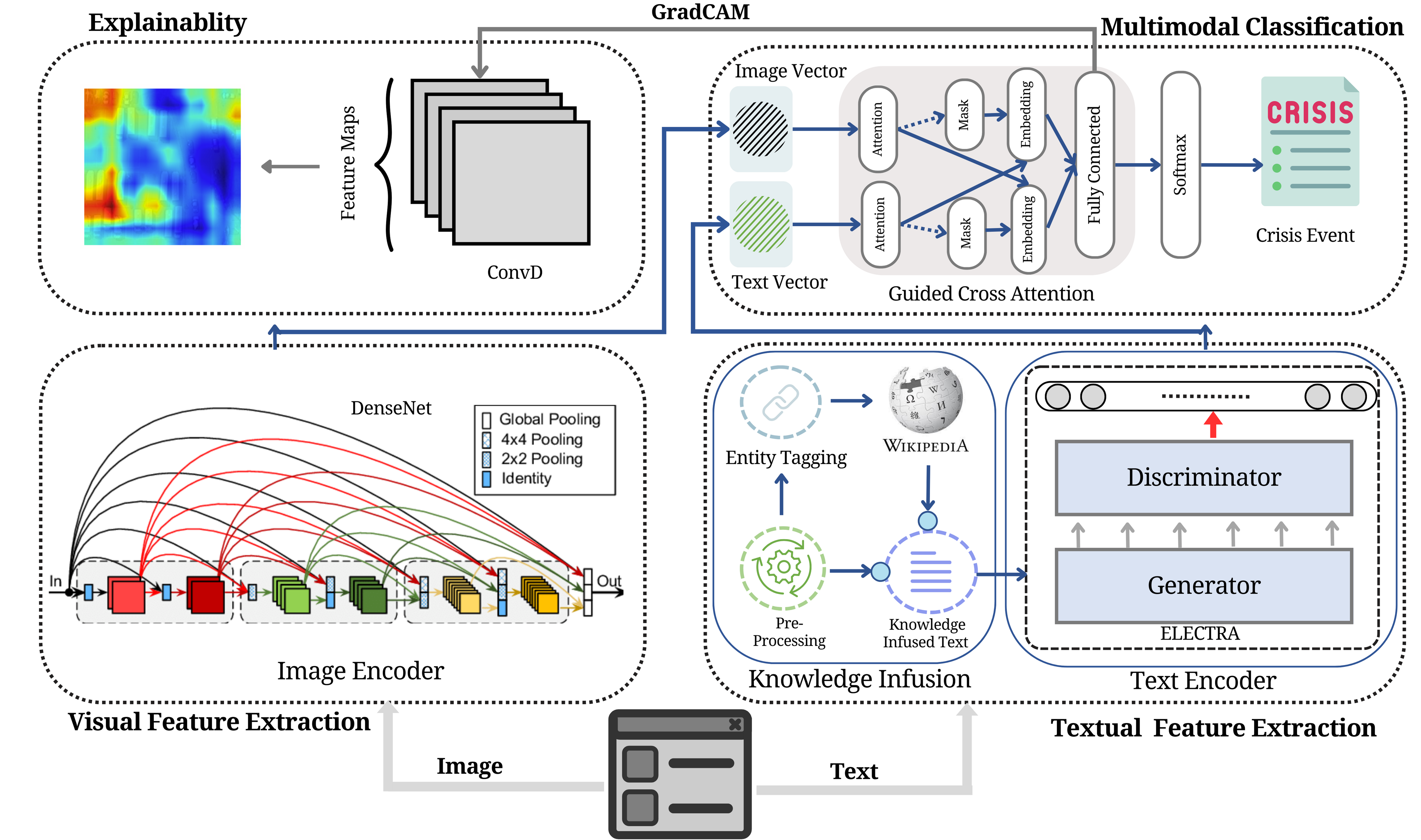}
    \caption{The overall architecture of CrisisKAN. }
    \label{fig:method}
\end{figure}

A brief overview of the proposed model, CrisisKAN is depicted in Fig. \ref{fig:method}. The model takes both text and image content as input and provide a probability distribution of labels over different event classes along with an attention mask for explainability. This framework comprises four key components: Visual feature extraction, Textual feature extraction, Guided Cross Attention Module, and Explainability. Description for each of these components is provided in the following sections.

\subsubsection{Visual Feature Extraction}
Convolutional Neural Networks (CNNs) is utilized to extract feature maps from images. We use DenseNet \cite{gao2017}, as it reduces module sizes while enhancing connections between layers. This addresses parameter redundancy and leads to improved accuracy. Although alternative methods like Ghostnet \cite{kai2020}, AlexNet \cite{alex2014} are viable, DenseNet is both efficient and a prevalent option for this purpose.
For each input image $I$, our approach yields the following outcome:
\begin{equation}
     \mathcal{Z}_I = \textrm{DenseNet}(I) \in \mathbb{R}^ {c \times h \times w}
\end{equation}
Here, $c$, $h$, and $w$ are the number of channels, height, and width of the feature vector respectively.


\subsubsection{Textual Feature Extraction}
Given a textual input denoted as $T$, comprising a set of words $\{w_1, w_2, \cdots, w_n\}$ associated with event-related information, text feature extraction involves two steps. First, it infuses text with external knowledge, followed by feeding this enhanced text into a transformer-based model to generate the final text-based feature vector.
\paragraph{Knowledge Infusion} 
To address challenges arising from biases inherent in events of similar types and the constraints of limited text, we incorporate external knowledge from Wikipedia using a proposed wiki extraction algorithm. We extract primary entities associated with the event from the given text. Relevant entities from external knowledge are identified using a relatedness score $\gamma_{rel}$ computed as follows for each word $w_i$ within the text $T$.
\begin{equation}
        \gamma_{rel} = Rel(w_i),\: w_i \in T
\end{equation}
Function $Rel(\cdot)$ provides relveancy of $w_i$ in Wikipedia. If $\gamma_{rel}$ is greater than some threshold then it is regarded as pertinent to the specific event and considered as primary entity denoted by $e$. Implementation details of $Rel(\cdot)$ and thereshold is provided in Section \ref{sec:data:set}. The assumption here is that text associated with the image contains the entities depicted in the image modality. Subsequent to the extraction of primary entities, we retrieve information about these entities from Wikipedia pages using $Wiki(\cdot)$ function and concatenate the gathered data. This process yields external information denoted as $T^{wiki}$.
\begin{equation}
        T^{wiki} = Concat([e_1^{wiki}, \cdots, e_k^{wiki}])
\end{equation}
The detailed algorithm to generate $T^{wiki}$ is presented in the Algorithm \ref{algo1}.

\begin{algorithm}[!ht]
\caption{Entity2Wiki Knowledge Extraction Algorithm} 
\textbf{Input: $T$, Input Text}\\ 
\textbf{Output: $T^{wiki}$, Wikipedia Text} 
\begin{algorithmic}[1]
	\ForEach{$w_i \in  T$}
	\State Calculate $\gamma_{rel}$ on $w_i$
    \If{$\gamma_{rel} > thres$}
	\State $T^{wiki}=T^{wiki} \cdot Wiki(w_i)$
	\EndIf
	\EndFor
    \State Return $T^{wiki}$
\end{algorithmic} 
\label{algo1}
\end{algorithm}
\paragraph{Text Encoder}
We utilized the ELECTRA \cite{clark2020} model to encode sequences of tokens in sentences. Although there are many transformer variants such as BERT \cite{devlin-etal-2019-BERT}, XLNET \cite{yang2019}, etc. are available but ELECTRA is considered more parameter-efficient and faster to train than BERT due to its replaced token detection objective and generator-discriminator setup. Once we collect the domain information in the form of external knowledge $T^{wiki}$ related to the event, we create final text for the input to  the final input for ELECTRA by fusing original text $T$ and external knowledge $T^{wiki}$ with a special ``[SEP]'' tag. Subsequently, these tokens were tokenized and input into the ELECTRA model to derive the final layer embedding $h^i$  in the following manner: 

\begin{align}
    S = Concat([T, T^{wiki}]),\: \: [h_S^0, h_S^1, \dots, h_S^{N_S}] = \textrm{ELECTRA}(S),  \mathcal{Z}_S = h_S^0
\end{align}
Here $\mathcal{Z}_S\in \mathbb{R}^ {n \times l \times d}$ and $n$, $l$, $d$ are batch size, maximum sequence length, dimension of feature vector.

\subsubsection{Guided Cross Attention-based Fusion}
We examined the image feature map $\mathcal{Z}_I$ and the text feature map $\mathcal{Z}_S$ derived from the encoding process in both modalities. Notable semantic inconsistencies are identified during training that impact the overall performance of the final model. For example, during tweet classification, a given modality might contain irrelevant or misleading content, potentially leading to unfavorable information exchange. We modified the cross attention module from \cite{abavisani2020multimodal} by incorporating self-attention as a preceding method. This provides guidance for the cross attention module. In other words, our model is meticulously structured to mitigate the influence of one modality on the other, employing guided self-attention as follows:
\begin{equation}
Attn(Q, K, V) = softmax\left(\frac{QK^T}{\sqrt{d}}\right)V
\end{equation}
Here $Q$, $K$, and $V$ are the query, key and value. $\sqrt{d}$  is a normalization factor that helps control the scale of the dot products and ensures more stable gradients during training. Now, new representations of $\mathcal{Z}_I$ and $\mathcal{Z}_S$ are achieved as follows:
\begin{align}
&    \overline{\mathcal{Z}_I} = Attn(\mathcal{Z}_I), \:\:\:
    \overline{\mathcal{Z}_S} = Attn(\mathcal{Z}_S)
\end{align}
Now, project new representation of image $\overline{\mathcal{Z}_I}$ and text $\overline{\mathcal{Z}_S}$ into a fixed dimensionality $K$ in following manner:
\begin{align}
& \tilde{\mathcal{Z}_I}=F\left(W_I^T \overline{\mathcal{Z}_I}+b_I\right), \:\:\: 
 \tilde{\mathcal{Z}_S}=F\left(W_S^T \overline{\mathcal{Z}_S}+b_S\right)
\end{align}
where $F$ represents an activation function such as ReLU and both $\tilde{\mathcal{Z}_I}$ and $\tilde{\mathcal{Z}_S}$ are of dimension $K$ and $K$ is fixed to 100.

Next to apply cross attention, attention masks $\alpha_{\mathcal{Z}_I}$ and $\alpha_{\mathcal{Z}_S}$ on  $\overline{\mathcal{Z}_I}$ and $\overline{\mathcal{Z}_S}$ are calculate as follows:

\begin{align}
& \alpha_{\mathcal{Z}_I}=\sigma\left({W_I^{\prime}}^T \overline{\mathcal{Z}_S}+b_I^{\prime}\right),\:
 \alpha_{\mathcal{Z}_S}=\sigma\left(W_S^{\prime T} \overline{\mathcal{Z}_I}+b_S^{\prime}\right)
\end{align}
where $\sigma$ is the Sigmoid function. $\alpha_{\mathcal{Z}_I}$ for the image is completely dependent on the text embedding $\overline{\mathcal{Z}_S}$, while the attention mask $\alpha_{\mathcal{Z}_S}$ for the text is completely dependent on the image embedding $\overline{\mathcal{Z}_I}$. Once the attention masks $\alpha_{\mathcal{Z}_I}$ and $\alpha_{\mathcal{Z}_S}$ for the image and text are obtained, we can enhance the projected image and text embeddings $\tilde{\mathcal{Z}_I}$ and $\tilde{\mathcal{Z}_S}$ by applying element-wise multiplication with $\alpha_{\mathcal{Z}_I} \cdot \tilde{\mathcal{Z}_I}$ and $\alpha_{\mathcal{Z}_S} \cdot \tilde{\mathcal{Z}_S}$. The final stage of this module involves taking the combined embedding, which represents both the image and text pair, and passing it into fully-connected network and the classification is performed using the conventional softmax cross-entropy loss.

\subsubsection{Explainability}
The primary motivation for incorporating an explainability module is to analyze errors in predictions and establish a transparent system through visual explanations. We employ explainablity in our CrisisKAN by using a model-specific method, known as Grad-CAM \cite{grad2017}. It is calculated using gradient information flowing into last convolution layer of the model. To employ this, we have added a new convolution layer (ConvD) to identify and visualize importance of each modality feature vector for building decision. In order to obtain, the class-discriminative localization map for a particular crisis event class label $y$, the method first computes the gradient of the score $G^{y}$ before the softmax with respect to $k$ number of feature maps $A^k$. To obtain grad weights ($\alpha_{k}^{y}$), these gradients are global average-pooled using in which ($w,h$) is the width and height dimensions and $Z$ is normalizing factor. In summary, $\frac{\partial G^y}{\partial A_{i j}^k} $ is gradients value by backpropogation and $\sum_{i=1}^{w} \sum_{j=1}^{h}$ is global average pooling in Eq \ref{eq:explain_2}. 
\begin{align}
&\alpha_{k}^{y} = \frac{1}{Z} \sum_{i=1}^{w} \sum_{j=1}^{h} \frac{\partial G^y}{\partial A_{i j}^k}, \: \:
    I_{\text {Grad-CAM }}^y=\textit{ReLU}\left(\sum_{k} \alpha_k^y \cdot A^k\right)\label{eq:explain_2}
\end{align}
Finally, Grad-CAM for image modality $I_{\text {Grad-CAM }}^y$ is obtained by ReLU activation function over weighted sum of feature maps which is shown in Eq. \ref{eq:explain_2}. As a result, Grad-CAM provides a class-specific heatmap to visualize important region in the image modality as a visual explanations.


\section{Experiments and Results}
\label{sec:exp}
In order to evaluate the efficacy of CrisisKAN, extensive experiments are conducted using the CrisisMMD dataset \cite{alam2018crisismmd}.  This section reports dataset, experimental settings, results and ablation study.

\subsection{Dataset \& Settings}
\label{sec:data:set}
CrisisMMD \cite{alam2018crisismmd} is widely recognised multimodal crisis dataset. This dataset was collected from the social media platform Twitter and contains images and textual information related to seven major crisis events including earthquakes, hurricanes, wildfires, and floods that happened in the year 2017. It is also categorized into three distinct tasks. Task 1 is designed to categorize whether a given image-text pair is informative and non-informative. On the contrary, Task 2 is concerned with categorising  the impact of event in five  classes, which include infrastructure damage, vehicle damage, rescue efforts, affected individuals (injury, dead, missing, found etc.), and others. Task 3 focuses on severity assessment, categorising as severe, mild, and little/no damage. We evaluate these tasks in two distinct settings similar to SOTA \cite{abavisani2020multimodal} method where  
Setting A considers only image-text pairs with identical labels, while Setting B incorporates all types of labeled image-text pairs for training, with test data matching that of Setting A. Table \ref{tab:dataset} shows the distribution of training, validation, and testing data for different tasks and settings.  

\begin{table}[]
\caption{Dataset distribution with different split, tasks and settings}
\centering
\resizebox{0.4\linewidth}{!}{%
\begin{tabular}{ccllll}
\hline
Setting            & \multicolumn{1}{l}{Task} & Train                     & Val                      & Test                     & Total                     \\ \hline
\multirow{3}{*}{A} & Task 1                                & \multicolumn{1}{c}{9601}  & \multicolumn{1}{c}{1573} & \multicolumn{1}{c}{1534} & \multicolumn{1}{c}{12708} \\
                   & Task 2                                & \multicolumn{1}{c}{2874}  & \multicolumn{1}{c}{477}  & \multicolumn{1}{c}{451}  & \multicolumn{1}{c}{3802}  \\
                   & Task 3                                & \multicolumn{1}{c}{2461}  & \multicolumn{1}{c}{529}  & \multicolumn{1}{c}{530}  & \multicolumn{1}{c}{3520}  \\ \hline
\multirow{2}{*}{B} & Task 1                                & \multicolumn{1}{c}{13608} & \multicolumn{1}{c}{1573}  & \multicolumn{1}{c}{1534} & \multicolumn{1}{c}{16715} \\
                   & Task 2                                & 8348                      & 477                      & 451                    & 9276                                     \\ \hline
\end{tabular}%
}

\label{tab:dataset}
\end{table}

\subsection{Experimental Setup \& Evaluation Metrics}
We trained our model across tasks and settings with the following parameters: (i) base learning rate: $2 \times 10^{-3}$, (ii) decay: 10X, (iii)  batch size: 64, (iv) optimizer: Adam with $\beta$1 = 0.9, $\beta$2 = 0.999, and $\epsilon$ = $1 \times 10^{-4}$, (v) loss function: categorical cross-entropy, (vi) epochs: 50. We have used $Tagme$ \cite{tagme2010} API as $Rel(\cdot)$ function to identify the relatedness of entities in the input text. Also, we have varied $thres$ from 0 to 0.5 and considered 0.1 threshold to label a word as relevant entity. These models are executed on NVIDIA A100 with 40 GB GPU memory. Symbols such as `@', `\#' and hyperlinks are removed as part of textual pre-processing. The model's performance is assessed using standard metrics such as classification accuracy, macro F1-score and weighted F1-score. These metrics show the performance of model with respect to each task.

\subsubsection{Multi-task Model Strength (MTMS)} The metrics defined above are assessed in relation to each task. Therefore, to evaluate the cumulative performance across all tasks, a new performance metric is introduced (Eq. \ref{eq:mtms}) called as Multi-task Model Strength (MTMS) which shows the overall strength of model. 
\begin{equation}
    \text{MTMS} = \sum_{i=1}^{i=n} \beta_{i} \times Acc_{i}
    \label{eq:mtms}
\end{equation}
Here $\beta_{i} = \frac{c_{i}}{\sum_{i=1}^{i=n} c_{i}}$, $Acc_{i}$ is accuracy and $c_{i}$ is number of classes for task $i$. The MTMS ranges from 0 to 1, with a value of 0 signifying that accuracy is zero across all tasks. MTMS assigns more importance to tasks with a higher number of classes, as the complexity of classification task increases in correlation with the number of classes they entail.

\begin{table*}[!t]
\caption{Comparative study on Setting A and B in terms of classification Accuracy (Acc)\%, Macro F1-score (M-F1)\%, Weighted F1-score (W-F1)\% and  Multi-task Model Strength (MTMS)\%.}
\centering
\resizebox{.9\linewidth}{!}{%

\begin{tabular}{llllllllllllll}
\hline
\multicolumn{1}{c}{\multirow{2}{*}{\textbf{Method}}} &
  \multicolumn{3}{c}{\textbf{Task 1}} &
   &
  \multicolumn{3}{c}{\textbf{Task 2}} &
   &
  \multicolumn{3}{c}{\textbf{Task 3}} &
   &
  \multicolumn{1}{c}{\multirow{2}{*}{\textbf{MTMS}}} \\ \cline{2-4} \cline{6-8} \cline{10-12}
\multicolumn{1}{c}{}     & Acc  & M-F1 & W-F1 &  & Acc  & M-F1 & W-F1 &  & Acc  & M-F1 & W-F1 &  & \multicolumn{1}{c}{} \\ \hline \hline
                         &      &      &      &  & \multicolumn{3}{c}{\textbf{Setting A}} & &  & & & \\ \hline
Unimodal DenseNet \cite{gao2017}        & 81.6 & 79.1 & 81.2 &  & 83.4 & 60.5 & 87.0 &  & 62.9 & 52.3 & 66.1 &  & 76.9                \\
Unimodal BERT  \cite{devlin-etal-2019-BERT}          & 84.9 & 81.2 & 83.3 &  & 86.1 & 66.8 & 87.8 &  & 68.2 & 45.0 & 61.1 &  & 80.5                \\
Unimodal ELECTRA \cite{clark2020}        & 86.3 & 82.9 & 85.5 &  & 87.2 & 67.4 & 88.2 &  & 68.8 & 49.1 & 62.4 &  & 81.5                 \\ \hline

Cross-attention \cite{abavisani2020multimodal}         & 88.4 & 87.6 & 88.7 &  & 90.0 & 67.8 & 90.2 &  & 72.9 & 60.1 & 69.7 &  & 84.5                \\
CentralNet  \cite{vielzeuf2018centralnet}             & 87.8 & 85.3 & 86.1 &  & 89.3 & 64.7 & 89.8 &  & 71.1 & 57.4 & 68.7 &  & 83.5                \\
GMU   \cite{arevalo2017gated}                   & 87.2 & 84.6 & 85.7 &  & 88.7 & 64.3 & 89.1 &  & 70.6 & 57.1 & 68.2 &  & 82.9               \\
CBP \cite{fukui2016cbpmultimodal}                     & 87.9 & 85.6 & 86.4 &  & 90.2 & 66.1 & 89.8 &  & 65.8 & 60.4 & 69.3 &  & 82.4                \\
CBGP  \cite{kiela2018efficient}                   & 88.1 & 86.7 & 87.3 &  & 84.7 & 65.1 & 88.7 &  & 67.9 & 50.7 & 64.6 &  & 80.3                \\
MMBT   \cite{kiela2019supervised}                  & 86.4 & 85.3 & 86.2 &  & 88.7 & 64.9 & 89.6 &  & 70.1 & 59.2 & 68.7 &  & 82.7                \\
VisualBERT  \cite{li2019visualBERT}             & 88.1 & 86.7 & 88.6 &  & 87.5 & 64.7 & 86.1 &  & 66.3 & 56.7 & 62.1 &  & 81.3                \\
PixelBERT    \cite{huang2020pixel}            & 88.7 & 86.4 & 87.1 &  & 89.1 & 66.5 & 88.9 &  & 65.2 & 57.3 & 63.7 &  & 81.8                \\
VilT     \cite{kim2021vilt}                & 87.6 & 85.1 & 88.0 &  & 86.7 & 61.2 & 87.2 &  & 67.6 & 58.4 & 65.0 &  & 81.2                \\ 
\textbf{CrisisKAN(Ours)} & \textbf{91.7} & \textbf{90.3}  & \textbf{91.2} &  & \textbf{93.6} &  \textbf{70.3} & \textbf{93.4} &  & \textbf{73.1}  &  \textbf{63.1}   &   \textbf{72.2}  &  &  \textbf{87.1}                    \\ \hline
 &      &      &      &  & \multicolumn{3}{c}{\textbf{Setting B}} & &  & & & \\ \hline

Unimodal DenseNet \cite{gao2017}        & 84.4 & 82.8 & 84.6 &  & 74.8 & 60.7 & 79.9 &  & - & - & - &  & 77.5 \\ 
Unimodal BERT \cite{devlin-etal-2019-BERT}           & 83.9  & 80.4 & 83.2 &  & 82.7 & 59.2 & 81.7 &  & - & - & - &  & 83.1\\
Unimodal ELECTRA \cite{clark2020}         & 84.6  & 81.1 & 83.8 &  & 83.4 & 60.1 & 83.5 &  & - & - & - &  & 83.8  \\ \hline
Cross Attention \cite{abavisani2020multimodal} & 85.6  & 82.3 & 84.8 &  & 89.3 & 63.4  & 89.8 &  & - & - & - & &88.2    \\ 
\textbf{CrisisKAN(ours)} & \textbf{86.9} & \textbf{84.5} & \textbf{86.2} &  &  \textbf{90.1}   &  \textbf{65.3 } & \textbf{90.9} &  & - & - & - & & \textbf{89.2 } \\ \hline
\end{tabular}%
}

\label{tab:setting_a}
\end{table*}

\subsection{Quantitative Results}
We evaluate our proposed methodology against several state-of-the-art technaiques including unimodal based and multimodal based methods. We compare our approach with single modality networks DenseNet for image, language models BERT and ELECTRA across the all tasks. The second category includes existing image-text multimodal classification methods \cite{abavisani2020multimodal,vielzeuf2018centralnet,arevalo2017gated,fukui2016cbpmultimodal,kiela2018efficient,kiela2019supervised,li2019visualBERT,huang2020pixel,kim2021vilt}. Some of the methods \cite{abavisani2020multimodal,li2019visualBERT,vielzeuf2018centralnet} emphasize on multimodal fusion based on global features obtained by each modality backbone, while other studies like \cite{fukui2016cbpmultimodal,kiela2018efficient} employ compact bilinear pooling based fusion. The quantitative results for these baselines are presented in Table \ref{tab:setting_a} for Setting A and B. We assess the efficacy of each baseline within the current dataset configuration, as shown in Table \ref{tab:dataset}. 

From Table \ref{tab:setting_a}, we can observe that all multimodal methods perform better than unimodal models, demonstrating the strength of multimodality learning. Moreover, compared with multimodal baselines, the accuracy of our CrisisKAN is significantly improved by approximately $3\%-5\%$ in Task 1, $3\%-7\%$ in Task 2 and $1\%-8\%$ in Task 3 with Setting A. We also calculate the multi task model strength (MTMS) across three tasks and can interpret that CrisisKAN achieves high MSMT with the score of $87.1\%$. 

In the Table \ref{tab:setting_a}, we also investigate our model with addition of noisy data in Setting B where image and text pair are labelled differently for same event. We find that CrisisKAN performs better compare to unimodal models and multimodal baseline both. Hence we can conclude from these results that  our model is effectively learning the textual and visual features together.

\subsection{Qualitative Results}
\begin{figure}
    \centering
    \includegraphics[width=.9\linewidth,height=5.4in]{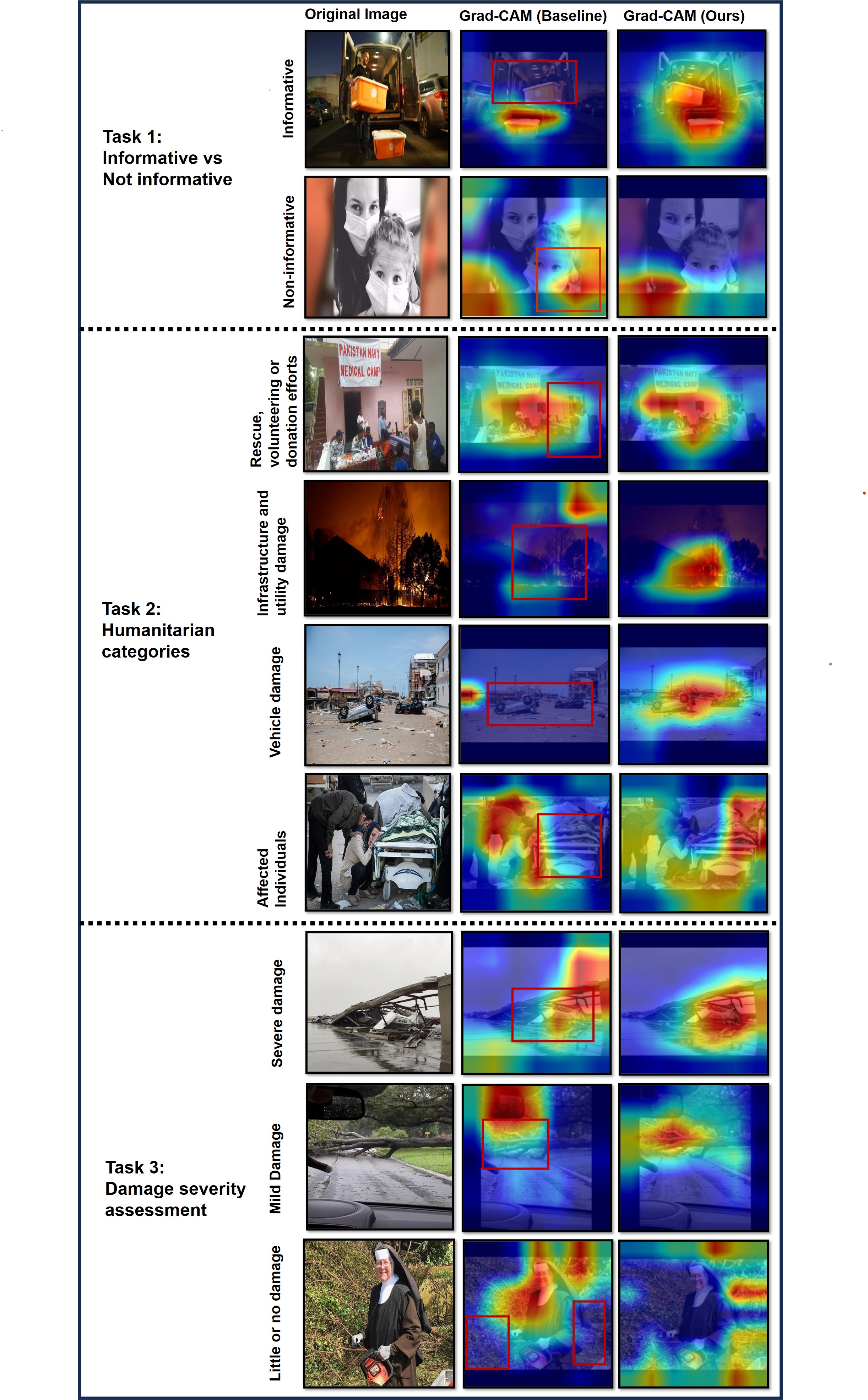}
    \caption{Comparative study of visual explanation for CrisisKAN (ours) with baseline model \cite{abavisani2020multimodal} across various tasks in Setting A.}
    \label{fig:explain_result}
\end{figure}

Along with accuracy, we also present our model's capabilities in terms of explainablity. The Fig. \ref{fig:explain_result} show the comparative study of visual explainations for baseline model \cite{abavisani2020multimodal} and ours model's prediction by highlighting the regions. We show the explainablity results across the three tasks where impact of features increase with the change of color gradient from blue to orange. The red box in the baseline model's results express those regions which is responsible for correct event classification and correctly learned by our proposed model. For task 1, our model's Grad-CAM map for informative class put more focus in the middle region and as by visual verification this result with the original image that regions are important to classify correctly in respective class. Similar way, the map for other tasks shows that effective and contributing features are extracted in the model leveraging guided cross attention. Finally, our visualizations on these examples take a step toward to build more trust on the proposed model.

\begin{figure}
  \begin{minipage}{0.5\linewidth}
  \vspace{-.1cm}
    \centering
    \includegraphics[width=0.7\linewidth]{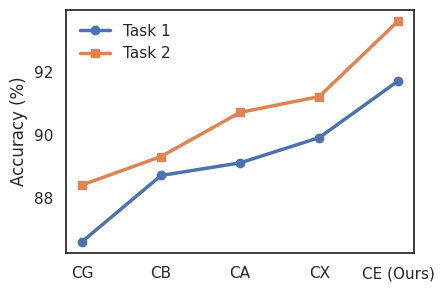}
    \caption{Comparison on different \\image and text encoders.}
    \label{fig:ab1}
  \end{minipage}%
  \begin{minipage}{0.5\linewidth}
  \vspace{-.2cm}
    \centering
    \includegraphics[width=.7\linewidth]{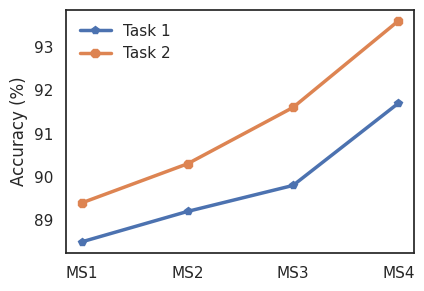}
    \caption{Comparison on different \\CrisisKAN Model Settings (MS).}
    \label{fig:ab2}
  \end{minipage}
\end{figure}

\subsection{Ablation Study}
\label{subsec:al}

\subsubsection{Leveraging different image and text encoders:}

To evaluate our model effectiveness, we leverage different image and text encoder such as CrisisKAN+ Ghostnet \textbf{(CG)}, CrisisKAN+BERT \textbf{(CB)}, CrisisKAN+ALBERT \textbf{(CA)}, CrisisKAN+XLNet \textbf{(CX)}, and CrisisKAN+ELECTRA \textbf{(CE)} (ours). This comparative analysis is conducted on Task 1 and Task 2 within Setting A, and the obtained results are depicted in Fig. \ref{fig:ab1}. The graph distinctly illustrates that ELECTRA outperforms the other models, achieving an accuracy of 91.7\% and 93.6\% for Task 1 and Task 2, respectively. Additionally, it's noteworthy that XLNet and ALBERT yield comparable outcomes, each with an accuracy of 89.9\% and 91.2\% for Task 1, and 89.1\% and 90.7\% for Task 2. This study establishes that through the integration of diverse image and text encoders, CrisisKAN significantly surpasses existing multimodal methodologies with significant margin.
\vspace{-0.8cm}
\subsubsection{Comparison among CrisisKAN variants:}
\label{subsubsec:al_setting}
Furthermore, we conduct a comparative analysis of the components within CrisisKAN \textbf{(MS4)} in relation to two specific aspects: the integration of knowledge and the utilization of guided attention mechanism. This study involves three distinct model settings: 1) without external knowledge and without guided attention \textbf{(MS1)}, 2) without Wikipedia knowledge \textbf{(MS2)}, and 3) without guided attention \textbf{(MS3)}. The corresponding outcomes are presented in Fig. \ref{fig:ab2} which provide the key insight that the inclusion of guided attention leads to an approximate 1\% increase over the performance of \textbf{(MS1)}. Additionally, the introduction of external knowledge results in improvements of around 2\% for both Task 1 and Task 2. Ultimately, when both guided attention and external knowledge are combined, ours \textbf{(MS4)} demonstrates a notable enhancement of 3-4\% over \textbf{(MS1)}, and 2-3\% over both \textbf{(MS2)} and \textbf{(MS3)}, for both tasks.

\section{Conclusion \& Future Scope}
\label{sec:conc}
\vspace{-0.1cm}
In our research, we introduce CrisisKAN, a novel Knowledge-infused and Explainable Multimodal Attention Network. This network is specifically designed to classify crisis events by intelligently combining features from both images and text. In order to address the challenge of event biases, we have integrated external knowledge sourced from Wikipedia into our model. To facilitate better communication among diverse modalities and to sift through irrelevant or potentially misleading data, we propose a guided cross attention module. This module effectively narrows down the semantic gap between modalities, enabling us to selectively fuse only the valuable information. Our research also delves into the realm of explainability. We have taken strides towards bridging the gap between the enigmatic ``black box'' models to models that can be explained and interpreted. To evaluate the effectiveness of our model, we evaluate our model on different crisis tasks and settings. Form the results, it is evident that our model not only outperforms many image and text only models but also other SOTA multimodal approaches.

While our proposed CrisisKAN model has a limitation due to the availability of external knowledge from Wikipedia only after an event has occurred, this drawback can be mitigated by extending the model to incorporate real-time information from social media during the event itself.  For the future direction, we are actively exploring additional ways for enhancing the performance of our image and external knowledge module. Also, we are committed to seamlessly integrating lengthy textual inputs into our model to further strengthen it capabilities. Furthermore, the explainability module will be expanded through the integration of text modality visualization, enabling the analysis of effects or errors associated with each modality. 

\bibliographystyle{splncs04}
\bibliography{main.bib}

\end{document}